\documentclass[10pt,twocolumn,letterpaper]{article}

\usepackage{iccv}

\usepackage{graphicx}
\usepackage{times}
\usepackage{amsmath}
\usepackage{amssymb}
\usepackage{epsfig}
\usepackage{booktabs}
\usepackage{multirow}
\usepackage{xcolor}

\usepackage[pagebackref,breaklinks,colorlinks]{hyperref}
\usepackage{caption}

\usepackage[capitalize]{cleveref}
\crefname{section}{Sec.}{Secs.}
\Crefname{section}{Section}{Sections}
\Crefname{table}{Table}{Tables}
\crefname{table}{Tab.}{Tabs.}

\iccvfinalcopy 

\def\iccvPaperID{5401} 

\ificcvfinal\fi

\begin{document}

\title{Compositional 3D Human-Object Neural Animation}



\author{Zhi Hou, Baosheng Yu, Dacheng Tao\\
The University of Sydney\\
{\tt\small \url{https://zhihou7.github.io/CHONA}} 
}

\twocolumn[{%
\renewcommand\twocolumn[1][]{#1}%
\maketitle
\begin{center}
    \centering
    \captionsetup{type=figure}
    \includegraphics[width=\linewidth]{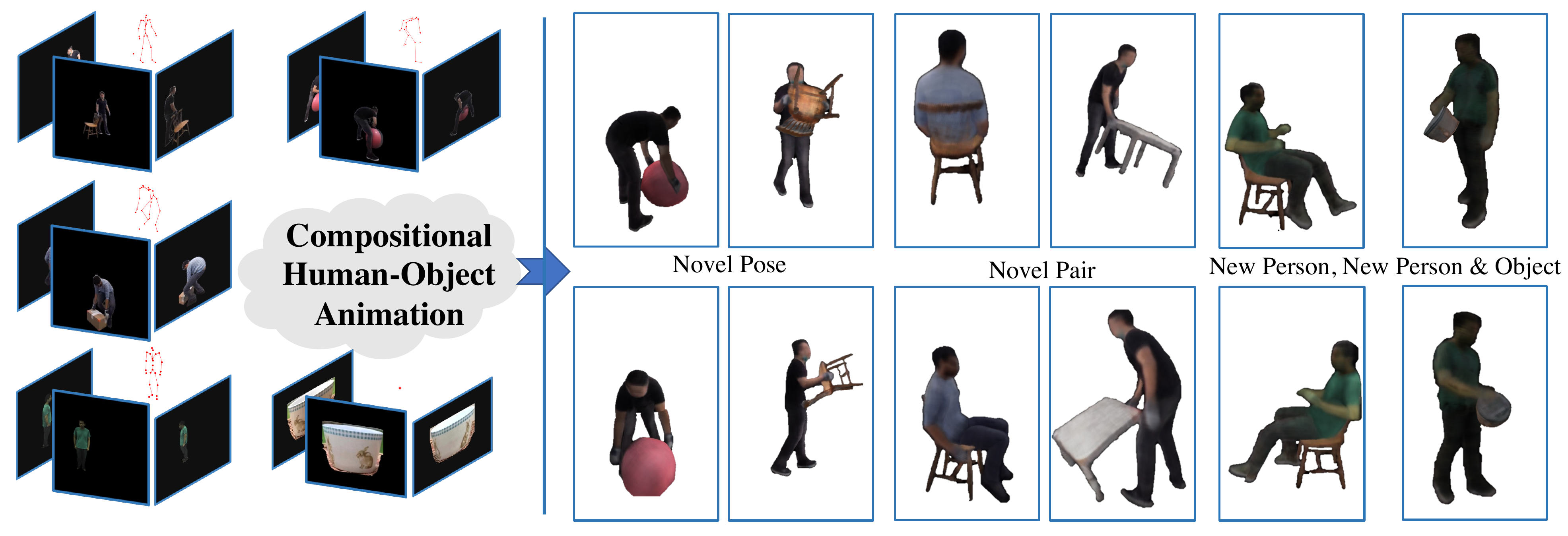}
    \captionof{figure}{An illustration of compositional human-object neural animation. Given a set of sparse multi-view RGB HOI short videos with less than 50 frames, we render the neural animation of novel HOIs with novel pose, human, and object. Specifically, most faces in the training dataset are partly blurred. 
    }
    \label{fig:demo}
\end{center}%
}]
\maketitle
\ificcvfinal\thispagestyle{empty}\fi

\begin{abstract}
Human-object interactions (HOIs) are crucial for human-centric scene understanding applications such as human-centric visual generation, AR/VR, and robotics. Since existing methods mainly explore capturing HOIs, rendering HOI remains less investigated. In this paper, we address this challenge in HOI animation from a compositional perspective, i.e., animating novel HOIs including novel interaction, novel human and/or novel object driven by a novel pose sequence. Specifically, we adopt neural human-object deformation to model and render HOI dynamics based on implicit neural representations. To enable the interaction pose transferring among different persons and objects, we then devise a new compositional conditional neural radiance field (or CC-NeRF), which decomposes the interdependence between human and object using latent codes to enable compositionally animation control of novel HOIs. Experiments show that the proposed method can generalize well to various novel HOI animation settings. Code will be made publicly available.
\end{abstract}

\section{Introduction}
\label{sec:intro}


Rendering 3D human-object animation is of great importance for human-centric generation with a wide range of real-world applications such as telepresence, video games, films, content generation, AR/VR and robotics. However, reconstructing and rendering human avatars with the interactive objects remains poorly investigated. Since traditional 3D reconstruction methods highly depend on dense cameras or depth sensors~\cite{schonberger2016structure,dou2016fusion4d,dou_motion2fusion_2017}, implicit neural representations for graphical objects present appealingly realistic results without the requirement of complex hardware and thus receive increasing attention from the community~\cite{mescheder2019occupancy,mildenhall2021nerf,barron_mipnerf_2022,niemeyer2021giraffe,yang2021learning}. Specifically, \cite{mildenhall2021nerf} introduces an implicit representation, i.e., neural radiance fields (NeRF), which represents static or rigid 3D objects/scenes as color and density fields and is capable of efficiently learning 3D geometry from images with differentiable volume rendering techniques. 



To explore dynamic non-rigid scenes and objects, vanilla NeRF has been recently extended to handle deforming scenes~\cite{park_nerfies_2021,tretschk_non-rigid_2021,peng_neural_2021,park2021hypernerf} and motion modeling~\cite{li_neural_2021,xian_space-time_2021,pumarola_d-nerf_2021}. Though the deformation-based approaches~\cite{park_nerfies_2021,park2021hypernerf,pumarola2021d} achieve appealing results in handling dynamic scenes, they are usually limited to scenes with small variances (\eg, faces). However, Human-centric interactions include large variances and extensive occlusions, which poses a significant challenge to directly model the dynamics with the interpolation methods~\cite{park_nerfies_2021,park2021hypernerf}. Recently, those neural linear skinning approaches represent multiple frames of human body with implicit neural representations under the control of skeleton~\cite{peng_animatable_2021,liu2021neural,su_-nerf_2021,kwon_neural_2021,xu_h-nerf_2021,noguchi2021neural,zheng_structured_2022,li2022tava,wang2022arah}, achieving considerable performance in free-viewpoint human avatar rendering and demonstrating good generalization to novel human poses~\cite{li2022tava,wang2022arah}. However, the above-mentioned methods usually focus on the animation of either individual human body or object, leaving the interactions between human and object poorly investigated. Particularly, the interactions are seriously self-occluded and it will hamper the modelling of human body. Meanwhile, several methods propose to explore the interactions between human and the surrounding objects or environments~\cite{taheri2020grab,hassan_populating_2021,guzov_human_2021,zhao_compositional_2022,dabral_gravity-aware_2021,zhang2020phosa,xu2021d3d,liu2022hoi4d}. Nevertheless, they mainly aim to reconstruct human/object shape and appearance rather than rendering animatable HOIs.



To jointly capture dynamic human body and objects with mutually occlusions, we thus generalize deformable neural radiance fields for human-object interaction with an additional point to indicate the objects. With a simple object modeling strategy, we can relieve the dependence on the object prior model (e.g. object meshes) and effectively improve the reconstruction of human and object under complex occlusions.
Specifically, the objects are regarded as disconnected joints in correspondence to the human body, and we then construct ``pseudo bones'' based on the object joint together with human body bones to model and control the dynamics of human-object interactions. By doing this, we extend the idea of animatable volumetric avatars to human-object interactions with non-linear pose-dependent deformation fields. Therefore, with the carefully designed canonical human-object pose, the generalized deformation fields and coordinate-based volumetric rendering, we can reconstruct and animate existing self-occluded HOIs. 
Particularly, we notice those Mesh or SDF-based methods~\cite{peng_animatable_2021,wang2022arah} usually fail to reconstruct the complex object without object prior model (Please refer to Section~\ref{sec:exp1}). Considering that mesh or prior model is not always available for novel objects, we thus adopt 3D human pose and 6-DoF object pose to guide the reconstruction and rendering from multi-view images, and simplify the control of HOI animation. We introduce the details of neural human-object deformation in Section~\ref{sec:method:deformation}.





Nevertheless, it is impractical to obtain all those interaction poses for the rare objects, which will limit the application of Neural Human-Object Animation in the real world. Considering people usually interact with similar objects in the same way, we thus introduce a compositional human-object animation challenge, which is not only related to novel poses/actions but also novel human and object. To enable the compositionally animate human-object interaction, we introduce compositional conditional neural radiance fields (CC-NeRF) with latent codes for human and object respectively. Specifically, to facilitate the transfer of the interactions among different humans/objects, we then decompose the human and object latent codes via the compositional invariant learning. By doing this, we thus enable the controllable animation for novel human-object interactions. We introduce the details of compositional animation in Section~\ref{sec:method:compositional}.



In this paper, we present a novel approach, named as compositional 3D human-object neural animation or CHONA, to implicitly reconstruct HOIs from sparse multi-view videos via coordinate-based neural representations, and compositionally animate HOIs under novel poses/interactions, novel person and novel object. An illustration of compositional HOI animation is shown in Figure~\ref{fig:demo}. Our contributions can be summarized as: 
\begin{itemize}
    \item We introduce an HOI animation framework by exploring neural HOI deformations.
    \item We devise a compositional conditional NeRF for compositional HOI animation, which enables transferring interaction poses to novel human and object.
    \item We perform comprehensive experiments to demonstrate that the proposed method not only improves the animation performance but also the compositional generalization.
\end{itemize}




\section{Related Work}

\subsection{Human-Object Interaction}
The interaction with objects is common in the people's daily life~\cite{jain_interactive_2009,gupta2009observing}. Early work mainly investigate synthesizing human pose and object~\cite{jain_interactive_2009},  human body reconstruction~\cite{fieraru2020three}, object recognition~\cite{wei2016modeling}, or human 3d pose estimation~\cite{andriluka2012human,kim2014shape2pose,chen2019holistic++,li2019estimating} under the interaction with objects or environments. Recently, Human-Scene Synthesis~\cite{zhang2020place,zhang2020generating,hassan2021populating,zhao2022compositional,wang2021synthesizing,hassan2021stochastic,wang2022towards} has attracted extensive interests from the community due to the potential applications in the content generation. Those methods usually depend on the prior human model (\eg, SMPL) and only synthesize the human motion in the scenes, while compositional Human-Object neural animation aims to animate both human and object in a compositional manner.
Recently, increasing approaches~\cite{bhatnagar2022behave,taheri2020grab,sun_neural_2021,haresh_articulated_2022,jiang_neuralhofusion_2022,dabral_gravity-aware_2021,xu2021d3d,zhang2020phosa,huang2022intercap,wang2022reconstructing,xie2022chore} focus on 3D Interactions between Human and its surrounding objects. Zhang~\cite{zhang2020phosa} present to reconstruct the spatial arrangements of Human-Object Interaction. ~\cite{xu2021d3d,dabral_gravity-aware_2021} reconstruct the meshes of human-object interactions, while recent work~\cite{jiang_neuralhofusion_2022} introduces the neural representations to human-object interaction and significantly advances the novel view synthesis performance. Particularly, a real HOI dataset, BEHAVE~\cite{bhatnagar2022behave}, consisting of 8 subjects and diverse objects, is introduced with spare views of HD videos and the poses of human and object. We mainly conduct our experiments based on BEHAVE. Concurrent work~\cite{haresh2022articulated,wang2022reconstructing,xie2022chore,huang2022intercap,zhang2022couch,xie2022chore,zhou2022toch} focus on reconstruction, 3D tracking or motion refinement, significantly ignoring interaction animations.
Besides, though current compositional approaches on human-centric interactions have studied the recognition~\cite{kato2018compositional}, detection~\cite{hou2020visual}, object affordance~\cite{hou2021affordance}, 2D generation~\cite{nawhal2020generating}, and 3D human-scene synthesis~\cite{zhao2022compositional}, the compositional 3D animation remains unsolved.

\subsection{Animatable Avatars}
3D Avatars~\cite{loper2015smpl,peng_neural_2021, wang2022arah,li2022tava,chen2021snarf,zhao2022humannerf} have been through a significant progress. Early work usually leverages SMPL~\cite{loper2015smpl} model to reconstruct or synthesize human body, however the body is usually naked. Recently, neural fields have dominated 3D shape representations and novel view synthesis. Peng~\etal~\cite{peng_neural_2021} present to implicitly reconstruct human body from spare videos with neural radiance fields with a carefully designed skinning deformation. Next, ~\cite{peng_animatable_2021,liu2021neural,kwon_neural_2021,noguchi2021neural,su_-nerf_2021,li2022avatarcap,zheng_structured_2022,su2022danbo} significantly facilitate the performance in novel view rendering for novel poses. More recently, ~\cite{li2022tava,wang2022arah} demonstrate appealing avatar generation under out of distribution poses. Meanwhile, ~\cite{li2022avatarcap,weng2022humannerf} presents to reconstruct high-fidelity human avatars from the monocular RGB video observation. Specifically, ~\cite{su2022danbo,li2022tava} requires only 3D skeletons and multiple multi-view frames to construct an animatable Avatar, while~\cite{wang2022arah} requires a pre-trained body SDF model to control the animation. Considering the fewer requirements on body models (\eg, SMPL) of pose-dependent animation, we follow~\cite{su2022danbo,li2022tava} to compositionally control the animation.

\subsection{Neural 3D Representations}
Neural representations~\cite{park2019deepsdf,mescheder2019occupancy,chen2019learning} have revolutionized the 3d surface representation, and achieved continuous, high resolution outputs of arbitrary shape. Recently, NeRF~\cite{mildenhall2021nerf} represents 3D points in the scene with density and color, and renders the scene with volumetric rendering techniques, achieving photorealistic novel rendering. Next, ~\cite{barron_mipnerf_2022} extends NeRF
to represent the scene at a continuously-valued scale with conical frustum. GRAF~\cite{schwarz2020graf} represents the neural radiance conditioned on shape/appearance latent codes. GIRAFFE~\cite{niemeyer2021giraffe} further presents controllable image synthesis with Compositional Generative Neural Feature Fields. However, those approaches~\cite{schwarz2020graf,niemeyer2021giraffe} mainly learn the representations from static scenes, in which the objects are not frequently occluded. Differently, Compositional Human-Object animation requires to control the synthesis from spare multi-view HOI videos, which includes massive occlusion for human body and objects. Meanwhile, \cite{schwarz2020graf,niemeyer2021giraffe,yang2021learning} mainly control the image synthesis for the static 3D scenes with rigid objects via linear transformation, while our approach is able to deform the interaction in a non-linear way.


\section{Method}
\label{sec:method}

In this section, we introduce the proposed compositional 3D human-object animation approach. Specifically, we first provide an overview of HOI animation and the popular neural radiance fields (NeRF). We then introduce the neural human-object deformation and the compositional conditional radiance fields in detail.

\subsection{Overview}

\begin{figure*}
  \centering
 \includegraphics[width=\linewidth]{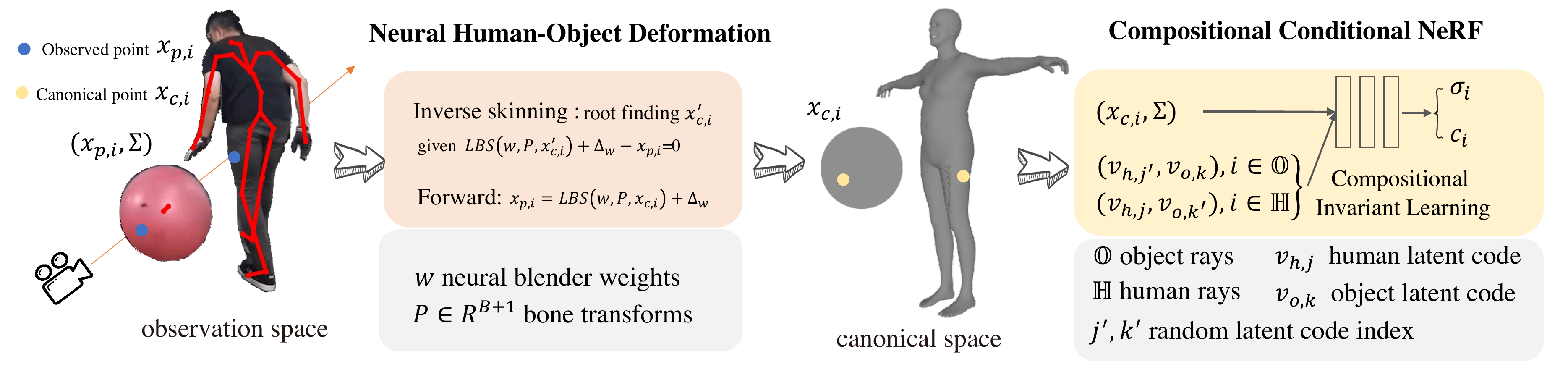}
 \caption{Overview of the proposed approach. The proposed compositional human-object neural animation approach leverages the neural Human-Object deformation module to deform the canonical points to posed points, and identify the corresponding canonical points of observed points via inverse skinning. Next, we obtain the density and color of the ray points conditioned on human and object latent codes, and accumulate the samples to render the pixel color. In addition, a compositional invariant learning strategy is introduced to decompose the interdependence between the two latent codes based on object and human masks, and facilitate compositional human-object animation.}
 \label{fig:overview}
\end{figure*}

Given sparse multi-view inputs, including interaction videos, single person videos, and objects, Compositional 3D Human-Object Animation enables to not only render the interaction under novel interaction pose, but also animate the interaction between a novel person and novel objects. Specifically, we build a pseudo bone for the object, and then treat the pseudo bone equally with body bones. Next, following the popular body skinning deformation techniques~\cite{peng_animatable_2021,zheng_structured_2022,li2022tava,wang2022arah}, we devise a neural Human-Object deformation method to construct animatable human-object interactions as illustrated in Figure~\ref{fig:overview}. Moreover, in order to control the interaction animation with novel people or objects, we devise compositional conditional radiance fields with two conditional latent codes, representing human and object respectively, to control the human-object identity. Specifically, we devise a compositional invariant learning strategy to decompose the interdependence between human and object latent codes, and thus enable to compositionally control the animation for novel objects or people.

\subsection{Preliminary: Neural Radiance Fields}
NeRF~\cite{mildenhall2021nerf} leads to significant progress in a wide range of 3D vision topics. It implicitly represents the geometry and appearance of the scene with a multi-layer perceptron neural network and volumetric rendering techniques. For a ray $\boldsymbol{r}$ and the viewing direction $(\theta, \phi)$, NeRF first queries the emitted color $c$ and density $\sigma$ at the 3D point $\boldsymbol{x}=(x,y,z)$ in the ray $\boldsymbol{r}$, then uses volumetric rendering to get the pixel color $C(\boldsymbol{r})$ via accumulating the view-dependent colors along the ray $\boldsymbol{r}$ as follow,

\begin{equation}
\begin{aligned}
C(\boldsymbol{r}) &= \sum_i^N T_i(1-exp(\sigma_i\delta_i))\boldsymbol{c}_i,
\end{aligned}
\end{equation}
where $\quad T_i = exp(-\sum_{j=1}^{i-1}\sigma_j\delta_j)$, $\delta_i$ indicates the distances between the sample points along the ray. Then, it optimizes the network via calculating the distance loss between $C(\boldsymbol{r})$ and ground truth pixel color. Recently, Mip-NeRF~\cite{barron_mipnerf_2022} extends NeRF via taking each point in the ray as a cone, the samples $\boldsymbol{x}$ along the ray as conical frusta modeled as multivariate Gaussians ($\mu$, $\Sigma$). Mip-NeRF accumulates the pixel color in a similar way to NeRF~\cite{mildenhall2021nerf}.

\subsection{Neural Human-Object Deformation}
\label{sec:method:deformation}

To represent the human-object interaction via neural volumetric representation, we devise a neural Human-Object deformation as follows. For each point $\boldsymbol{x}_p$ in the observed/posed space, we have a corresponding point $\boldsymbol{x}_c$ in the canonical space, which can be deformed into $\boldsymbol{x}_p$ via neural skinning deformation. The canonical representation includes a Lambertian neural radiance field $F_{\Theta_{\boldsymbol{r}}}:(\boldsymbol{x}_c, \boldsymbol{\Sigma}) \to (\boldsymbol{c}, \boldsymbol{\delta})$, where $r$ denotes the pixel ray, $\boldsymbol{c}=(r, g, b)$ indicates the material color, $\boldsymbol{\delta}$ represents the material density
respectively. Besides, we follow Mip-NeRF~\cite{barron_mipnerf_2022} to accumulate the
samples $(\boldsymbol{x}_p,\boldsymbol{\Sigma})$ (a multivariate Gaussian) to render the pixel color at each ray.
As illustrated in Figure~\ref{fig:overview}, the canonical human-object space includes a T-pose body and an object placed in front of the body. We denote the transformation of each body bone as $\boldsymbol{T}_i, 0\leq i < B$ , where $B$ is the number of body bones and $\boldsymbol{T}_{i}\in R^{4\times4}$. We represent the 6 DoF transformation $(\boldsymbol{t}, \boldsymbol{r})$ of the object from canonical space to the observed space as $\boldsymbol{T}_{B}$. As a result, we have $\boldsymbol{P} = \{\boldsymbol{T}_0,\boldsymbol{T}_1,..., \boldsymbol{T}_B\}\in R^{(B+1)\times 4 \times 4}$ representing the transformation of a human-object interaction.\\

\noindent{\bf Forward Skinning.} Following the popular animatable avatar methods~\cite{chen2021snarf,wang_metaavatar_2021,wang2022arah,li2022tava}, we revise the traditional linear blend skinning (LBS)~\cite{anguelov2005scape,loper2015smpl,osman2020star} into neural skinning to deform a canonical Human-Object pose according to rigid bone transformations. We treat background as an additional bone. Thus, we have $B+2$ bones for the human-object interaction. Similar to~\cite{peng_animatable_2021,wang2022arah,li2022tava}, a MLP function $F_{\Theta_s}:\boldsymbol{x_c} \to \boldsymbol{w}$ is used to project a canonical point $\boldsymbol{x_c}$ into corresponding weights $\boldsymbol{w}$.  Given the skinning weights $\boldsymbol{w}=(w_0,w_1,...,w_{B-1},w_{B},w_{bg}) \in R^{B+2}$ and a pose $\boldsymbol{P} = \{\boldsymbol{T}_0,\boldsymbol{T}_1,...,\boldsymbol{T}_B\}$, we use forward LBS to define the deformation of a sample $\boldsymbol{x}_c$ in the canonical space to $\boldsymbol{x}_p$ in the view space:
\begin{equation}
  \begin{aligned}
  \boldsymbol{x}_p &= LBS(F_{\Theta_s}(\boldsymbol{x}_c), \boldsymbol{P}, \boldsymbol{x}_c)+F_{\Theta_\nabla}(\boldsymbol{x}_c, \boldsymbol{P})\\
   &= \lbrack \sum_{j=0}^{B+1} F_{\Theta_s, j}(\boldsymbol{x}_c) \cdot \boldsymbol{T}_j + w_{bg} \cdot \boldsymbol{I} \rbrack \cdot \boldsymbol{x}_c + F_{\Theta_\Delta}(\boldsymbol{x}_c, \boldsymbol{P}),
\end{aligned}
\end{equation}
where $\boldsymbol{I}\in R^{4\times4}$ denotes identity matrix, $F_{\Theta_\Delta}: (\boldsymbol{x}_c, \boldsymbol{P}) \to \Delta_w \in R^{3}$ is for modeling the non-linear deformations~\cite{li2022tava}. \\


%
\noindent{\bf Inverse Skinning.} It requires to transformer the observed points into canonical space for rendering the model. Similar to~\cite{chen2021snarf,wang2022arah,li2022tava}, we leverage the root finding strategy~\cite{chen2021snarf} to deform $\boldsymbol{x}_p$ to $\boldsymbol{x}_c^{'}$ subject to,
\begin{equation}
  \begin{aligned}
  f(\boldsymbol{x}_c^{’}) &= LBS(F_{\Theta_s}(\boldsymbol{x}_c^{'}), \boldsymbol{P}, \boldsymbol{x}_c^{'}) + F_{\Theta_\Delta}(\boldsymbol{x}_c^{'}, \boldsymbol{P}) - \boldsymbol{x}_p
  = 0,
\end{aligned}
\end{equation}
where $\boldsymbol{x}_c^{'}$ denotes the potential canonical point of $\boldsymbol{x}_p$. Then we solve it numerically via Newton’s method similar to~\cite{wang2022arah,li2022tava}, and we simply use the $K=5$ nearest bones of the point in observed space to initialize the Newton's method for reducing the computational burden.
We get $K$ corresponding points $\{\boldsymbol{x}_{c,0}^{'}, \boldsymbol{x}_{c,1}^{'}, ...,\boldsymbol{x}_{c,K-1}^{'}\}$ for the observed point $\boldsymbol{x}_p$. With the canonical points, we follow~\cite{chen2021snarf,li2022tava} to compute the gradients and render the image (See more details in Appendix).

\subsection{Compositional Conditional Radiance Fields}
\label{sec:method:compositional}

Though the proposed human-object neural deformation enables the animation for a given human-object interaction, it fails to animate novel combinations, \ie an interaction involves a novel person or a novel object. Due to the combinatorial explosion of Human-Object interactions and the challenges of capturing interaction poses, we can not collect the multi-view videos for all possible interactions, which significantly limits the potential applications of the proposed human-object neural deformation. Therefore, we devise Compositional Conditional Radiance Fields to enable compositionally animating the interactions from novel combinations, and even novel person and objects. To decouple the controlling of human and object, we use two latent codes for the Conditional Radiance Fields. Denote $\boldsymbol{v}_h\in R^{N_h \times C}$ and $\boldsymbol{v}_o\in R^{N_o \times C}$, where $N_h$ and $N_o$ are the numbers of person and object categories, as the latent codes of human and object respectively, we have the conditional radiance fields as follows,
\begin{equation}
  \label{eq:conditional}
F_{\Theta_{\boldsymbol{r}}}:(\boldsymbol{x}_c, \boldsymbol{\Sigma}, \boldsymbol{v}_{h,j}, \boldsymbol{v}_{o,k}) \to (\boldsymbol{c}, \boldsymbol{\delta}),
\end{equation}
where $0\leq j<N_h$ and $0\leq k<N_o$ denote the corresponding person and object for observed interaction. With the conditional radiance fields, we can control the rendering for different human-object pairs with $\boldsymbol{v}_h$ and $\boldsymbol{v}_o$. However, the two latent codes $\boldsymbol{v}_h$ and $\boldsymbol{v}_o$ are entangled together, \ie, the rendering is controlled jointly by two latent codes. Therefore, the conditional radiance field in Eq.~\eqref{eq:conditional} fails to animate the interactions of novel human or objects.\\

\noindent{\bf Compositional Invariant Learning.} To decouple the interdependence between the human and object latent codes, we further introduce a compositional invariant learning (CIL) strategy for conditional radiance fields, named as Compositional Conditional Radiance Fields, to ease the spurious correlation between human and object latent codes, and thus enable compositional neural animation. Specifically, for the pixel in human body, we expect ($\boldsymbol{c}, \boldsymbol{\delta}$) only dependent on $\boldsymbol{v}_h$, regardless of the value of $\boldsymbol{v}_o$. Thus, when the ray $\boldsymbol{r}$ is located in the human body, we randomly set the value for the object latent codes, and vice versa for rays in the object. Then, the color and density of the Compositional Conditional Radiance Fields for the points in ray $\boldsymbol{r}$ are presented as follows,

\begin{equation}
  \boldsymbol{c}, \boldsymbol{\delta} =\left\{
  \begin{array}{rcl}
   F_{\Theta_{\boldsymbol{r}}}(\boldsymbol{x}_c, \boldsymbol{\Sigma}, \boldsymbol{v}_{h,j^{'}}, \boldsymbol{v}_{o,k}) & & \boldsymbol{r} \in \mathbb{O} \\
   F_{\Theta_{\boldsymbol{r}}}(\boldsymbol{x}_c, \boldsymbol{\Sigma}, \boldsymbol{v}_{h,j}, \boldsymbol{v}_{o,k^{'}}) & & \boldsymbol{r} \in \mathbb{H} \\
\end{array}\right. \\
\end{equation}
where $ \mathbb{H} \cap \mathbb{O}=\emptyset $, $\mathbb{H}$ and $\mathbb{O}$ represent the rays set of human and object respectively. $j^{'}$ and $k^{'}$ are random latent human and object codes respectively. Via randomly setting the latent codes, we can decouple the interdependence of the Human-Object pairs in the training set. Therefore, we can control the animation via human or object latent codes individually.

\section{Implementation Details}
In this paper, we follow the linear blender skinning deformation~\cite{peng_animatable_2021,wang2022arah,zheng_structured_2022,li2022tava} to devise an additional pseudo bone for human-object interaction and two latent codes for compositional animation. We localize the human and object rays according to the provided segmentation in~\cite{bhatnagar2022behave}. We model the shading effect similar to~\cite{li2022tava}, and sample 64 points along a ray. Due to the camera distance difference between different datasets, we sample 2048 rays for BEHAVE images, 1024 rays for ZJU-mocap images, and 512 rays for CO3D images in each mini-batch. Meanwhile, we utilize two additional losses for skinning weights and non-linear deformations in~\cite{li2022tava} for optimization. The overall loss function thus is $\mathcal{L}=\mathcal{L}_{img} + \lambda \mathcal{L}_w+\beta \mathcal{L}_\Delta$, where $\mathcal{L}_{img}$ indicates image loss similar to~\cite{mildenhall2021nerf}, $\mathcal{L}_w$ represents the loss to encourage the onehot skinning weights $\boldsymbol{w}$, $\mathcal{L}_\Delta$ is to encourage the non-linear deformation term $F_{\Theta_\Delta}(\boldsymbol{x}_c, \boldsymbol{P})$ close to zero. Both $\mathcal{L}_w$ and $\mathcal{L}_\Delta$ are MSE losses. In our experiments, we set $\lambda$ to 1.0 and $\beta$ to 0.1. The Adam optimizer~\cite{kingma2014adam} is adopted for the training with an initial learning rate 5e-4 and an exponentially decay strategy to 5e-6.

\begin{table*}
\begin{center}
\caption{Human-Object Animation under novel interactions. ARAH~\cite{wang2022arah}$*$ indicates we apply the neural human-object deformation to ARAH~\cite{wang2022arah}.
  }
  \label{tab:pose_animation_acts}
\resizebox{\linewidth}{!}{
\begin{tabular}{@{}lcc|cc|cc|cc|cc|cc|cc@{}}
\hline
\multirow{2}{*}{Method} & \multicolumn{2}{c}{backpack}&\multicolumn{2}{c}{chairblack}&\multicolumn{2}{c}{chairwood}&\multicolumn{2}{c}{suitcase}&\multicolumn{2}{c}{tablesmall}&\multicolumn{2}{c}{tablesquare}&\multicolumn{2}{c}{yogaball}\cr\cline{2-15}


&PNSR&SSIM &PNSR  & SSIM   &PNSR  & SSIM   &PNSR  & SSIM   &PNSR  & SSIM   &PNSR  & SSIM &PNSR  & SSIM  \cr
\hline
TAVA~\cite{li2022tava}& 27.9 & 0.960 & 28.3 & 0.959 & 26.0 & 0.960 & 28.9 & 0.964 & 25.7 & 0.965 & 22.8 & 0.943 & 24.6 & 0.950\\
ARAH~\cite{wang2022arah} & 27.9 & 0.969 & 28.4 & 0.970 &25.9 & 0.972& 29.3&0.971 & 25.5 & 0.976 & 22.8 & 0.966 & 24.9 & 0.960 \\
ARAH~\cite{wang2022arah}$*$ & 27.4 & 0.971 & 27.3 & 0.967 & 24.6 & 0.959  & 28.8 & 0.978 & 25.2 & 0.973 & 24.7 & 0.968 & 27.2 & 0.976 \\
Baseline & 27.8 & 0.969 & 28.3 & 0.969 & 25.9 & 0.963 & 28.8 & 0.971 & 26.1 & 0.967 & 26.3 & 0.963 & 28.2 & 0.973\\
CHONA (ours) &  {\bf28.4} & {\bf0.971} &{\bf 29.1 }&{\bf 0.971} &{\bf 27.3} &{\bf 0.969} &{\bf  29.4} &{\bf 0.974 }& {\bf 27.9} & {\bf 0.974 }&{\bf 27.9 }&{\bf 0.966} & {\bf28.2 }&{\bf 0.974}\\


\hline
\end{tabular}}
\end{center}
\end{table*}
\section{Experiments}

In this section, we perform experiments to quantitatively and qualitatively illustrate Human-object Animation, including novel poses, compositional animation, and even novel non-interactive person/static object, for the proposed method. More experiments (\eg, the effect of pose quality, the analysis of geometry) can be found on animation in Appendix. \\

\noindent\textbf{Datasets and Metrics.} We leverage three datasets, \ie, BEHAVE~\cite{bhatnagar2022behave}, ZJU-mocap~\cite{peng_neural_2021} and CO3D~\cite{reizenstein21co3d} for the compositional human-object neural animation. BEHAVE~\cite{bhatnagar2022behave} is a 4D dataset with 8 subjects performing a wide range of interactions with 20 common objects from 4 camera views. BEHAVE provides estimated body poses and object poses for each interaction, while each interaction has less than 50 frames. BEHAVE includes many blurred faces and frames, which we provide the analysis in supplementary materials. ZJU-mocap~\cite{peng_neural_2021} consists of 10 sequences captured with 23 calibrated cameras. We select one subject (386) and four cameras for evaluating compositional HOI animation on novel persons. CO3D is a large 3D object dataset with multiple sequences. We select "bowl" for evaluation on novel objects. More objects are illustrated in Appendix. We adopt the popular metrics in animatable avatars, peak signal-to-noise ratio (PSNR) and structural similarity index (SSIM).


\subsection{Novel Pose Animation}
\label{sec:exp1}
In order to evaluate the novel pose animation, we select the first subject (S01), four kinds of different boxes, and seven classes of objects with distinct interactions from the BEHAVE dataset to construct a benchmark for novel pose animation. The objects with distinct interactions, including ``backpack'', ``chairwood'', ``chairblack'', ``suitcase'', ``tablesmall'', ``tablesquare'' and ``yogaball'' are utilized to evaluate novel actions animation. The boxes consist of four scales, \ie, ``boxtiny'',``boxsmall'',``boxmedium'' and ``boxlarge'', and we leverage it to demonstrate the effectiveness of the proposed method on different scales of objects. We randomly split the training and validation set for boxes, while we randomly choose two or one interaction in other classes for training and the remaining one for validation. The details can be found in supplementary materials. For each interaction in BEHAVE~\cite{bhatnagar2022behave}, there are only less than 50 frames. Therefore, we use one V100 GPU to run our experiments with 100,000 iterations.\\

\noindent\textbf{Methods.} For a robust comparison, we use template-free model TAVA~\cite{li2022tava} and model-based method ARAH~\cite{wang2022arah} as our baseline methods. Meanwhile, we apply the neural human-object deformation to ARAH for demonstration and further devise a baseline method with only object localization control. The details of those methods can be found in Appendix.\\

\noindent\textbf{Quantitative Comparisons.} Table~\ref{tab:pose_animation_acts} illustrates the proposed method considerably improves the baseline Methods among different objects. Without object modelling, TAVA~\cite{li2022tava} and ARAH~\cite{wang2022arah} fail to render the novel interactions. Particularly, if we apply the neural human-object deformation to the model-based Avatar method (\ie, ARAH~\cite{wang2022arah}), the model can reconstruct the simple object, \eg, ``yogaball'', but achieve worse results for the complex objects (\eg, ``chairblack'', ``chairwood''). We think it is challenging to reconstruct implicit object shapes for ray tracing in ARAH. Table~\ref{tab:pose_animation_boxes} demonstrates that the proposed method consistently improves the baseline methods on boxes, and the larger the box is, the better the performance of the proposed method is.  Visualized comparison in Appendix shows the small object, \eg, ``boxsmall'', only occupies a small region in the HOI images. Therefore, for those interactions, PNSR and SSIM can not well-demonstrate the model performance.\\


\begin{figure*}
  \centering
 \includegraphics[width=0.99\linewidth]{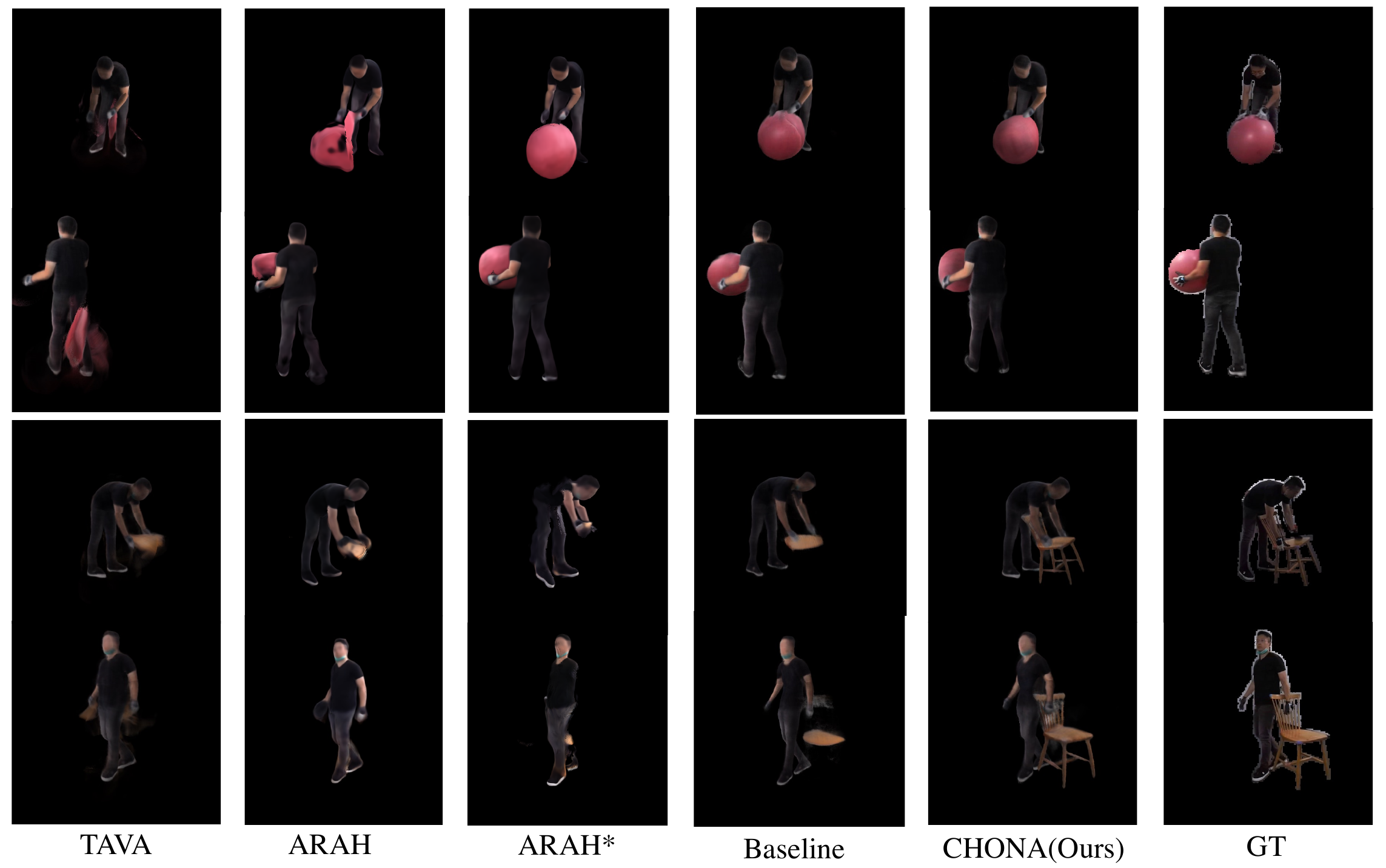}
 \caption{Visualized Comparisons between the proposed method, baseline methods (TAVA~\cite{li2022tava}, ARAH~\cite{wang2022arah}). We demonstrate the results of ``yogaball'' and ``chairwood'' with two distinct views. Video demos for better comparison are also provided in the supplementary materials.}
 \label{fig:comp_vis1}
\end{figure*}

\noindent\textbf{Qualitative Comparisons.} Figure~\ref{fig:comp_vis1} shows our method can effectively animate the human-object interactions under the control of poses. Without the modeling of objects, the baseline methods achieve poor performance on object rendering though it can still render the human body correctly. ARAH with neural human-object deformation (ARAH$*$) can reconstruct the ``yogaball'', but fail to reconstruct the ``chairwood'', and even fail to animate the body. We think the self-occlusion of HOI and the failure in reconstructing the complex object cause the model did not achieve a good human-object deformation generalization.

\begin{table}
\setlength\tabcolsep{1.5pt}
\begin{center}
\caption{Human-Object Animation for the boxes, \ie, different sizes of objects.
  }
  \label{tab:pose_animation_boxes}
\small

\begin{tabular}{@{}lcc|cc|cc|cc@{}}
\hline
\multirow{2}{*}{Method} & \multicolumn{2}{c}{boxlarge}&\multicolumn{2}{c}{boxmedium}&\multicolumn{2}{c}{boxsmall}&\multicolumn{2}{c}{boxtiny}\cr\cline{2-9}

&PNSR  & SSIM&PNSR  & SSIM&PNSR  & SSIM&PNSR  & SSIM \cr
\hline
TAVA~\cite{li2022tava}& 22.6 & 0.949 & 25.9 & 0.967 & 26.8 & 0.970 & 27.5 & 0.973 \\
ARAH~\cite{wang2022arah} & 23.3 & 0.963  & 26.3& 0.972 & 27.0 & 0.974 & 27.7 & 0.977 \\
CHONA & {\bf 27.2} & {\bf0.971} & {\bf28.5} & {\bf 0.976} &{\bf 28.0} & 0.974 &{\bf 28.3} & 0.976 \\



\hline
\end{tabular}
\end{center}
\end{table}

\subsection{Compositional Animation}
\label{sec:comp}
In this subsection, we provide experiments to evaluate the proposed method on the compositional Human-Object Animation. We first construct a compositional benchmark for compositional animation with two subjects (S01, S02) and nine objects, totally 18 combinations, from BEHAVE to construct a sub-dataset. Human-Object Interaction is composed of person, action, and object. There are at most three actions in BEHAVE. Therefore, the subset includes 37 combinations of $<person, action, object>$. Given a person, there are different novel compositions as follows,
\begin{itemize}
  \item Novel Action, \ie, the combination of the object and the person exists in the training set, but the action is novel. This is similar to novel pose animation.
  \item Novel Object, \ie, there are no combinations of the person and the object in the training set, but the combination of the action and the object exists in the training set.
  \item Novel Action and Object, \ie, the combination of the action and the object does not exist in the training set.
\end{itemize}
We then split the dataset into a training set and three validation sets, \ie, novel action validation, novel object validation and novel action-object validation. For similar objects, we treat the actions with the same name equally. For example, the action ``sit'' between ``chairwood'' and ``chairblack'' is treated equally. Then, we randomly select 13 (around 1/3) combinations as the training set, and split the remaining combinations into three novel categories according to the description above. There are 614 frames in the training set, we thus use two V100 GPUs to run our experiments with 300,000 iterations.\\

\begin{table}
\centering
\caption{Compositional Human-Object Animation. The subscripts of $_{a}$, $_{o}$ and $_{ao}$ indicate the results of novel action set, novel object set, novel action-object set respectively. CIL indicates compositional invariant learning.}
\label{tab:novel_obj}
\resizebox{\linewidth}{!}{
\begin{tabular}{@{}lcccccc@{}}
\hline
Method &PSNR$_{a}$ & SSIM$_{a}$ &PSNR$_{o}$  & SSIM$_{o}$ &PSNR$_{ao}$  & SSIM$_{ao}$ \cr
\hline
w/o CIL &  28.2 & 0.967 & 26.5 & 0.961 & 27.4 & 0.964 \\
CC-NeRF & 28.1 & 0.966 & {\bf 27.0} & {\bf 0.966} & {\bf 28.0}  & {\bf 0.968}\\
\hline
\end{tabular}}
\end{table}

\begin{figure}[!ht]
  \centering
 \includegraphics[width=\linewidth]{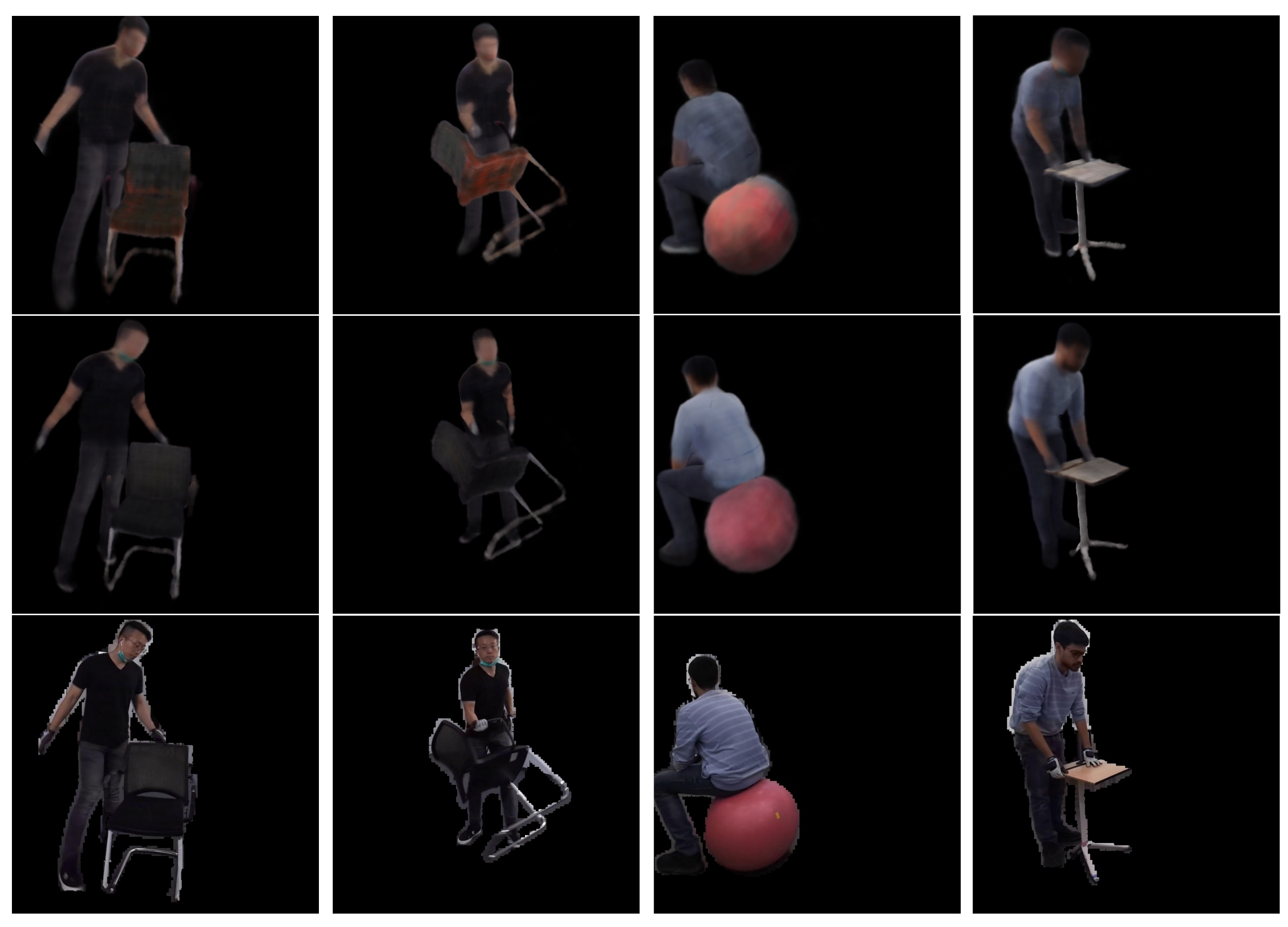}
 \caption{Visual comparisons between the proposed CC-NeRF and the baseline method (w/o CIL). The first row is the baseline, the second row is the proposed method, and the last row is ground truth. The first two columns indicate novel object categories, and the last two columns show novel action-object categories.}
 \label{fig:comp_vis2}
\end{figure}

\noindent\textbf{Comparisons.} The proposed CC-NeRF effectively improves the rendering for both human and object as illustrated in Table~\ref{tab:novel_obj}. We notice CC-NeRF achieves similar performance to the network without compositional invariant learning on novel action/pose animation. However, for the novel object split and novel action-object split, The proposed method effectively decomposes the control of different people and objects, and thus illustrates better performance on the compositional animation. Figure~\ref{fig:comp_vis2} demonstrates that the head (the mask is missing) in the baseline is dissimilar to the ground truth, but more similar to another subject. The objects of the baseline become red due to the entangling of the two latent codes. Those cases indicate CC-NeRF effectively decomposes the latent codes and achieves compositional animation.

\begin{figure}[!ht]
  \centering
 \includegraphics[width=\linewidth]{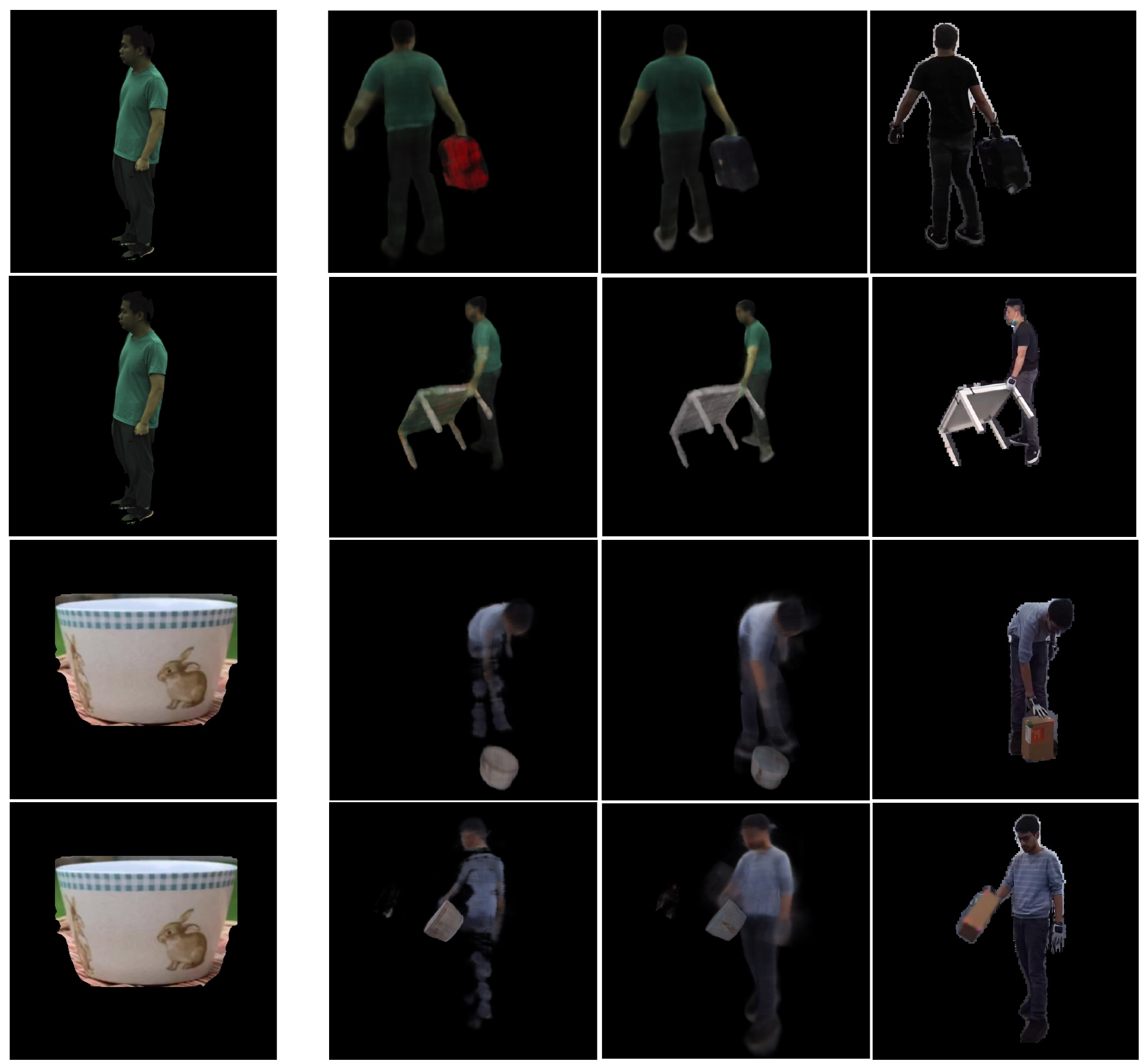}
 \caption{Visual comparisons between the proposed CC-NeRF and the baseline (w/o CIL) on novel object and person animation. The first column is novel/object, the second column is the baseline method, the third column is the proposed method, and the last column is the given pose. The first three rows indicate novel person animation, while the last two rows show novel object animation.}
 \label{fig:comp_vis3}
\end{figure}

\begin{table}[!ht]
\begin{center}
\caption{Comparison on novel static objects (``chairwood''). CIL indicates compositional invariant learning.
  }

\label{tab:novel_obj1}
\small
\begin{tabular}{@{}l|cccc@{}}
\hline

 Method & PNSR$_{ind}$ & SSIM$_{ind}$ & PNSR$_{ood}$ & SSIM$_{ood}$\cr
\hline


w/o CIL & 25.1 & 0.944 & 27.1 & 0.960 \\
CC-NeRF   & {\bf 27.4} & {\bf 0.956} & {\bf 29.2} & {\bf 0.965}\\


\hline
\end{tabular}
\end{center}
\end{table}

\subsection{Novel Person and Static Object}
The proposed method is not only effective for compositional animation, but also useful for transferring the interactions to non-interactive person and static object, which is more challenging and requires to effectively disentangle the interdependence among person, object and poses. We leverage the person (386) in ZJU-mocap and object bowl in CO3D to jointly train with BEHAVE. Here, for the dataset in BEHAVE, we directly adopt the subset in Section~\ref{sec:comp}. We observe the Compositional Conditional NeRF significantly improves the animation for novel object and person as illustrated in Figure~\ref{fig:comp_vis3}. We find the method without CIL will incur wired color on the object (\eg, the suitcase) or render the color of the novel person into the object (\eg, the ``tablesquare'' is green). Besides, for the novel object, we observe the baseline fails to render the human body. More visualized demonstration with additional objects and persons is provided in Appendix.
Table~\ref{tab:novel_obj1} further quantitatively show the superior performance on novel static object animation. Here, we select only one frame from ``chairwood'' and two action sets of ``chairblack'' for training. We split the validation set into ``ind'' (similar actions) and ''ood'' ( novel actions). The quantitative comparison for novel person is provided in Appendix.

\section{Conclusion and Future Work}
In this paper, we address the challenge of compositional human-object animation via neural Human-Object skinning deformations and compositional conditional radiance fields. Specifically, we construct a pseudo bone for the object, and devise a human-object skinning deforming approach to model the interactions between human and object.
Moreover, to enable compositional Human-Object animation, we further present compositional conditional neural radiance fields, which decompose the human and object latent codes via compositional invariant learning, to compositionally control the animation for novel human-object combinations, and even novel person and objects. Comprehensive experiments demonstrate the proposed method significantly improves the animation performance, as well as the compositional generalization.
Though we achieve considerable performance with the proposed methods, there still are some challenges, \eg, how to understand the interaction region (\ie, affordance region of the object). As human interacts with similar objects in a similar way, we think we can make use of the similarity of affordances among similar objects for the affordance localization to novel objects in the future.

Besides, we devise a compositional animation approach to transfer the interaction poses among similar objects to ease the challenge to capture interaction poses. However, it might be more beneficial to generate the object poses from the human motion poses according to the corresponding interaction categories since there are considerable collected human action poses, which we leave to future work.

{\small
\bibliographystyle{ieee_fullname}
\bibliography{egbib,3d_hoi}
}

\appendix
\section{Benchmark Construction}
\label{sec:data}
In our experiment, we randomly choose one action as validation set for each subject-object pair to evaluate the performance on out-of-distribution poses.

Here we provide the interactions splits of our experiments in Table~\ref{tab:splits}. For boxes, we randomly split the frames into training set and validation set because there is only a single interaction for each box. For compositional animation, we select ``yogaball'', ``chairblack'', ``chairwood'',``tablesquare'', ``tablesmall'', ``suitcase'', ``boxmedium'', ``boxlarge'', ``boxsmall'' from BEHAVE~\cite{bhatnagar2022behave} to construct the benchmark. Table~\ref{tab:splits2} present the splits of compositional animation.

\begin{table*}
\begin{center}
\caption{Dataset splits for novel pose animation.
  }
  \label{tab:splits}
\small

\begin{tabular}{@{}|l|c|c|@{}}
\hline

Objects & training set  & validation set \cr
\hline
backpack &  Date01\_Sub01\_backpack\_hug,Date01\_Sub01\_backpack\_back & Date01\_Sub01\_backpack\_hand \\
chairwood &  Date01\_Sub01\_chairwood\_hand,Date01\_Sub01\_chairwood\_sit & Date01\_Sub01\_chairwood\_lift \\
chairblack & Date01\_Sub01\_chairblack\_lift,Date01\_Sub01\_chairblack\_hand&Date01\_Sub01\_chairblack\_sit \\
suitcase & Date01\_Sub01\_suitcase\_lift & Date01\_Sub01\_suitcase \\
tablesmall & Date01\_Sub01\_tablesmall\_lift,Date01\_Sub01\_tablesmall\_move & Date01\_Sub01\_tablesmall\_lean\\
tablesquare & Date01\_Sub01\_tablesquare\_hand,Date01\_Sub01\_tablesquare\_lift & Date01\_Sub01\_tablesquare\_sit \\
yogaball & Date01\_Sub01\_yogaball & Date01\_Sub01\_yogaball\_play \\

\hline
\end{tabular}
\end{center}
\end{table*}

\begin{table*}
\begin{center}
\caption{Dataset splits for compositional animation.
  }
  \label{tab:splits2}
\small

\begin{tabular}{@{}|p{4cm}|p{4cm}|p{4cm}|p{4cm}|@{}}
\hline

 training set  & novel action validation & novel object validation & novel action object validation \cr
\hline
Sub01\_chairwood\_hand, Sub01\_chairwood\_lift,  Sub01\_tablesmall\_lean, Sub01\_tablesmall\_lift,  Sub01\_yogaball\_play, Sub02\_boxmedium\_hand,
     Sub02\_boxsmall\_hand, Sub02\_chairblack\_hand, 
     Sub02\_chairblack\_lift, Sub02\_suitcase\_ground, Sub02\_tablesquare\_sit, Sub02\_tablesquare\_lift,  Sub01\_boxlarge\_hand & Sub01\_yogaball, Sub02\_suitcase\_lift,  Sub01\_chairwood\_sit, Sub02\_chairblack\_sit, Sub01\_tablesmall\_move, Sub02\_tablesquare\_move, Sub01\_suitcase & Sub01\_chairblack\_sit, Sub02\_chairwood\_sit, Sub01\_suitcase\_lift, Sub02\_yogaball\_sit, Sub02\_tablesmall\_move, Sub01\_tablesquare\_hand & Sub01\_chairblack\_hand, Sub01\_chairblack\_lift, Sub02\_chairwood\_hand, Sub02\_yogaball\_play, Sub02\_tablesmall\_lean, Sub02\_tablesmall\_lift, Sub01\_tablesquare\_sit, Sub01\_tablesquare\_lift, Sub02\_boxlarge\_hand, Sub01\_boxmedium\_hand, Sub01\_boxsmall\_hand \\

\hline
\end{tabular}
\end{center}
\end{table*}

\section{Challenges Analysis on BEHAVE}
\label{sec:challenge}

{\bf Occlusions} BEHAVE~\cite{bhatnagar2022behave} is a real-world 3D HOI dataset with \textit{ only four camera views and extensive occlusions}, which poses a significant challenge for the detailed reconstruction of Human and Object. Meanwhile, each interaction has less than 50 frames, which is challenging for the model to implicitly reconstruct the human body and object.

\begin{figure*}
  \centering
 \includegraphics[width=\linewidth]{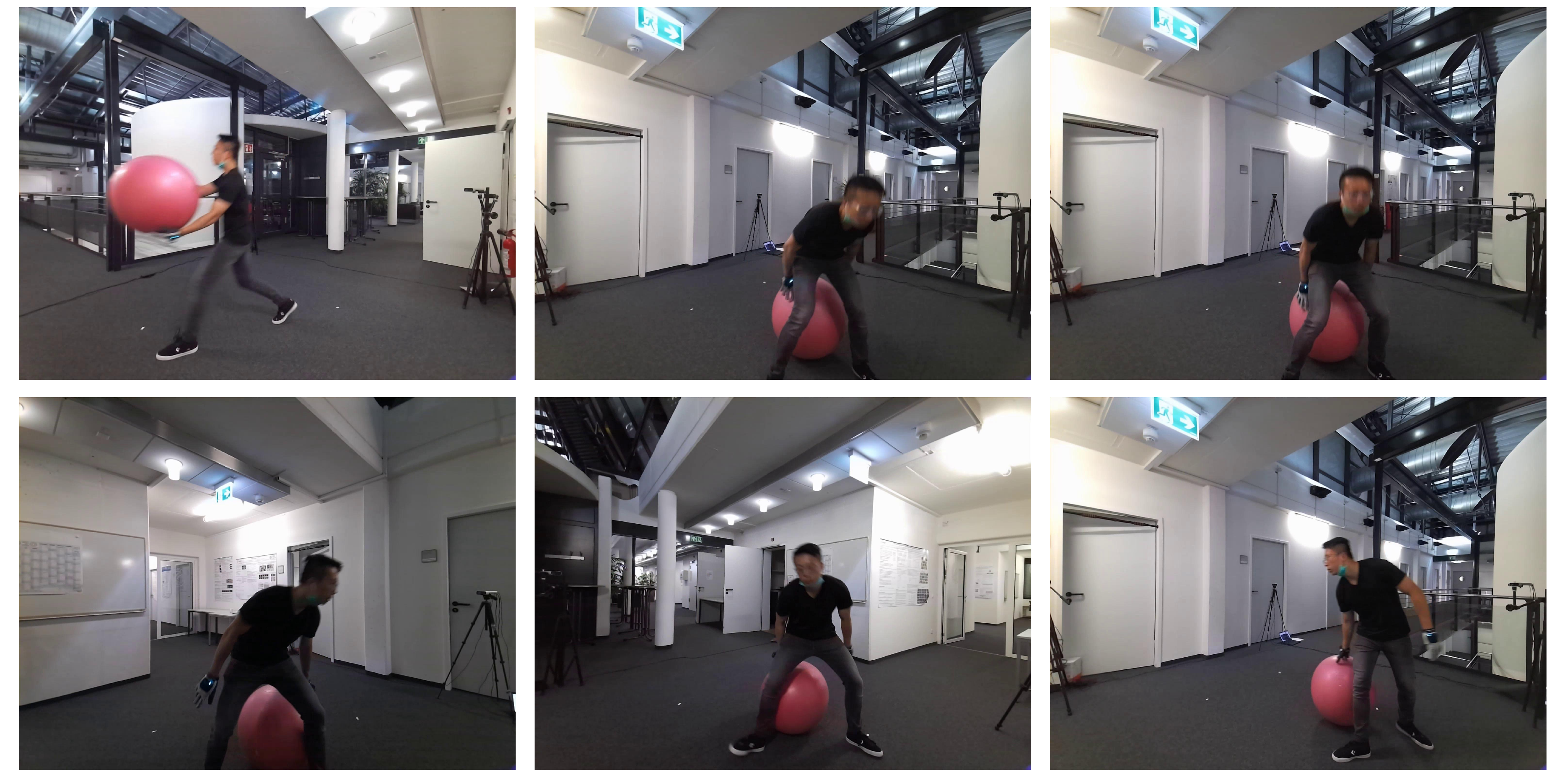}
 \caption{Illustration of the blurry faces and frames.}
 \label{fig:supp_data_face}
\end{figure*}

{\bf Blurry faces and frames} To protect privacy, BEHAVE~\cite{bhatnagar2022behave} uses the mask or fuzzy technique to blur most of the faces as illustrated in Figure~\ref{fig:supp_data_face}. This makes it very difficult to reconstruct the face of Subject01. Meanwhile, there are also blurry frames in BEHAVE. This further poses a significant challenge for a detailed reconstruction of HOI as illustrated in Figure~\ref{fig:supp_data_face}.

\begin{figure*}
  \centering
 \includegraphics[width=0.8\linewidth]{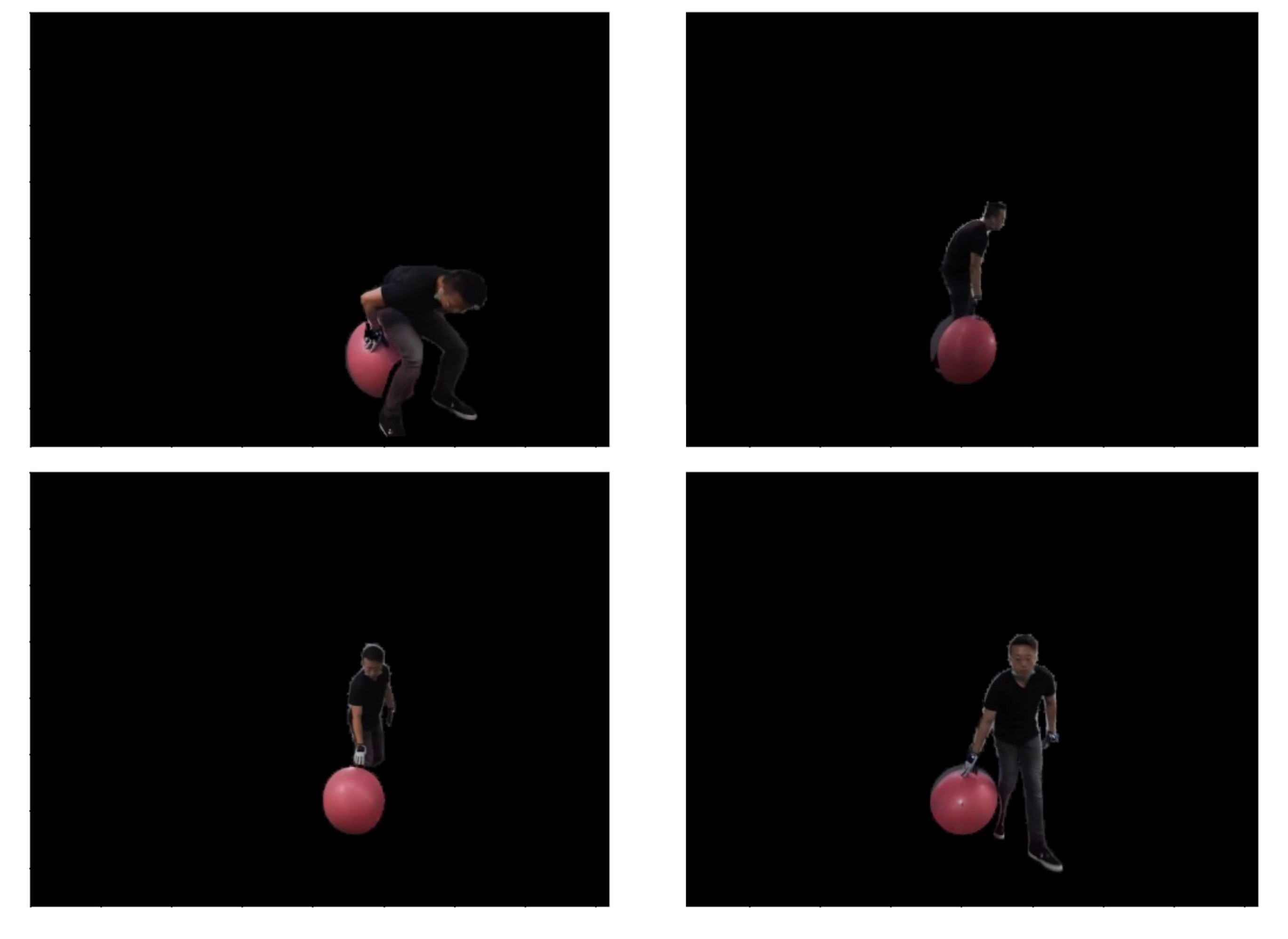}
 \caption{Illustration of inaccurate masks. The boundary between the yogaball and human is not correct. The wrong boundary even causes the shape of yogaball changes. }
 \label{fig:supp_data_mask_blurry}
\end{figure*}

{\bf Inaccurate Segmentation} Besides, the segmentation mask in BEHAVE~\cite{bhatnagar2022behave} is not much accurate due to the occlusion and complex background as illustrated in Figure~\ref{fig:supp_data_mask_blurry}. One can find more inaccurate segmentation in the ground truth of the video comparison. In our experiment, we find the proposed method is able to marginally implicitly reconstruct the object and human body. However, the segmentation problem also degrades the accuracy of reconstruction.

\end{document}